\documentclass{ieeeaccess}
\usepackage{cite}
\usepackage{amsmath,amssymb,amsfonts}
\usepackage{algorithmic}
\usepackage{graphicx}
\usepackage{textcomp}
\usepackage{verbatim}
\usepackage{stfloats}
\usepackage{bm}
\def\BibTeX{{\rm B\kern-.05em{\sc i\kern-.025em b}\kern-.08em
    T\kern-.1667em\lower.7ex\hbox{E}\kern-.125emX}}
\begin{document}
\history{Date of publication 4 January 2024, date of current version 11 January 2024.}
\doi{10.1109/ACCESS.2024.3349961}

\title{Multi-Task Convolutional Neural Network for Image Aesthetic Assessment}
\author{\uppercase{Derya Soydaner}\authorrefmark{1}, 
\uppercase{Johan Wagemans}\authorrefmark{1}}

\address[1]{Department of Brain and Cognition, University of Leuven (KU Leuven), Leuven, Belgium }
%\address[2]{Department of Brain and Cognition, University of Leuven (KU Leuven), Leuven, Belgium }

\tfootnote{This work was supported by the European Research Council (ERC) for the ‘Gestalts Relate Aesthetic Preferences to Perceptual Analysis’ (GRAPPA) project under Grant 101053925.
}

\markboth
{D. Soydaner, J. Wagemans: Multi-task convolutional neural network for image aesthetic assessment}
{D. Soydaner, J. Wagemans: Multi-task convolutional neural network for image aesthetic assessment}

\corresp{Corresponding author: Derya Soydaner (derya.soydaner@kuleuven.be)}

\begin{abstract}
As people's aesthetic preferences for images are far from understood, image aesthetic assessment is a challenging artificial intelligence task. The range of factors underlying this task is almost unlimited, but we know that some aesthetic attributes affect those preferences. In this study, we present a multi-task convolutional neural network that takes into account these attributes. The proposed neural network jointly learns the attributes along with the overall aesthetic scores of images. This multi-task learning framework allows for effective generalization through the utilization of shared representations. Our experiments demonstrate that the proposed method outperforms the state-of-the-art approaches in predicting overall aesthetic scores for images in one  benchmark of image aesthetics. We achieve near-human performance in terms of overall aesthetic scores when considering the Spearman's rank correlations. Moreover, our model pioneers the application of multi-tasking in another benchmark, serving as a new baseline for future research. Notably, our approach achieves this performance while using fewer parameters compared to existing multi-task neural networks in the literature, and consequently makes our method more efficient in terms of computational complexity. 

\end{abstract}


\titlepgskip=-21pt

\maketitle

\section{Introduction}
\label{sec:introduction}
\PARstart{I}{mage} aesthetic assessment is a challenging task due to its subjective nature. Some people may find an image aesthetically pleasing, while others may disagree. Aesthetic preferences of individuals are diverse and they can depend on many factors. Because of the importance and complexity of the problem, the literature on automated image aesthetic assessment is extensive \cite{deng2017, zhai2020}. In recent years, deep learning has become an important part of this literature based on its substantial impact in many areas. Given that deep neural networks can already perform tasks that were previously thought to be exclusive to humans, such as playing games \cite{schrittwieser2020}, it is not unreasonable to expect them to be able to assess the aesthetic value of images as well. Currently, image aesthetic assessment has a significant impact on many application areas such as automatic photo editing and image retrieval.   

In this context, neural networks have become a powerful tool in \emph{computational aesthetics}. This interdisciplinary field of research is of great importance for the automatic assessment of image aesthetics, and has led to the development of several state-of-the-art models for aesthetics research. In this study, we aim to evaluate a computational approach for image aesthetics which considers the overall aesthetic score as well as individual attributes that can impact aesthetic preferences. Therefore, we focus on the multi-task setting to assess the model's performance across multiple tasks. We handle the image aesthetic assessment task as a regression problem, i.e., our aim is predicting the aesthetic ratings for images. However, predicting overall aesthetic scores using regression-based approaches is complicated because aesthetic liking is influenced by a multitude of interacting factors. Many of these factors are subjective, and their combined effect is notoriously difficult to predict. This difficulty is compounded further in a multi-task setting, making the task even more challenging.  

To this end, we propose a multi-task convolutional neural network (CNN) that predicts an overall aesthetic score for a given image while also learning important attributes related to aesthetics. To demonstrate the effectiveness of our approach, we evaluate our multi-task CNN on two benchmark datasets in aesthetics research, namely the Aesthetics with Attributes Database (AADB) \cite{kong2016} and the Explainable Visual Aesthetics (EVA) dataset \cite{kang2020}. These datasets are unique in that they provide both overall aesthetic scores and attribute scores, making them valuable resources for evaluating image aesthetic assessment models. 

Our proposed multi-task CNN performs well for image aesthetic assessment, while being efficient in terms of computational complexity. Our multi-task CNN is the first of its kind applied to the EVA dataset, making it a new baseline for the multi-task setting on this dataset. Moreover, it achieves near-human performance on the overall aesthetic scores of the AADB dataset while having fewer parameters than the previous studies in the literature, demonstrating the principle of Occam's razor in machine learning.

The rest of the paper is organized as follows. In Section \ref{related-work}, related work is presented. We introduce our multi-task CNN in Section \ref{nn}. We describe our experimental setup in Section \ref{setup} and we discuss our results in Section \ref{results}. Finally, the conclusions and outlook are given in Section \ref{conclude}.

\subsection{Contributions}
Our main contributions are summarized as follows.

\begin{itemize}

\item We propose an end-to-end multi-task CNN for image aesthetic assessment and conduct systematic evaluation of our model on two image aesthetic benchmarks.   

\item In the multi-task setting, our model achieves the state-of-the-art result on the overall aesthetic scores of the AADB dataset, while requiring fewer parameters than previous approaches.

\item On the more recent EVA dataset, we conduct performance analysis and our model is the first multi-task CNN for this dataset, serving as the new baseline. 

\item Our evaluation shows that the multi-task setting consistently outperforms the single-task setting for the same neural network architecture across both datasets. 

\item As a result, we present a simple yet effective multi-task neural network architecture for image aesthetic assessment and provide a detailed evaluation of it on both image aesthetic datasets. 
\end{itemize}

\subsection{Problem formulation}
In this study, our aim is to develop a model that predicts aesthetic-related scores of images. We use aesthetic benchmarks that include images with overall aesthetic scores and scores for $K$ aesthetic attributes. Our model learns from the training set of $N$ samples $D = \left\{ (x^{(i)}, y^{(i)}) \right\}_{i=1}^{N}$. Here, each training sample consists of an RGB image $x^{(i)} \in \mathbb{R}^d$. Correspondingly, $y^{(i)} \in \mathbb{R}^{K+1}$ is a concatenated vector of the overall aesthetic score $y_o^{(i)} \in \mathbb{R}$ and scores for  $K$ aesthetic attributes $y_a^{(i)} \in \mathbb{R}^K$. Our model learns from this training data to accurately predict the aesthetic-related scores of images.

Such problems, where the output is a numerical value, are known as \textit{regression} problems. Here, the task is to learn the mapping from the input to the output. To this end, we assume a machine learning model of the form 

\begin{equation}
y = f(x|\theta),
\label{function}
\end{equation}
where $f(.)$ denotes the model and $\theta$ represents its parameters. Since we have images as input data, we choose the CNN as the model $f(.)$. We use a CNN in a \textit{multi-task} setting, as the target vector $y^{(i)}$ includes overall aesthetic score and scores for $K$ aesthetic attributes. Our objective is to obtain a network $f: \mathbb{R}^d \rightarrow \mathbb{R}^{K+1}$,  where $f$ can simultaneously predict the overall aesthetic scores and attribute scores from input images. 

\section{Related Work}
\label{related-work}
The task of image aesthetic assessment is typically approached as either a binary classification problem, where the aim is to classify an image as low or high aesthetics, or as a regression problem, where a model predicts an aesthetic score for a given image. Prior studies have investigated both classification and regression-based approaches to image aesthetic assessment. 
Deep learning techniques have achieved remarkable success in various fields, and image aesthetic assessment is no exception, as evidenced by the increasing number of studies exploring the use of deep neural networks in this area. It is clear that these techniques have played a critical role in contributing to notable advances in image aesthetics research. For example, Kang \emph{et al.} (2014) \cite{kang2014} presented a CNN that predicts image quality, while Lu \emph{et al.} (2014) \cite{lu2014} proposed an aesthetics classification network that learns several style attributes. Lee \emph{et al.} (2019) \cite{lee2019} utilized a Siamese network-based approach in this area. In another study, Lu \emph{et al.} (2015) \cite{lu2015} developed a deep neural network for image style recognition, aesthetic quality categorization, and image quality estimation. A neural network classifier assesses the aesthetic quality of an image, and this model can be implemented for contrast enhancement and image cropping \cite{lee2019a}. In some studies, models based on CNNs were used to predict a single aesthetic score for an image \cite{kao2015,kao2016}.  Inspired by a visual neuroscience model, Wang \emph{et al.} (2017) \cite{wang2017} introduced a  model for image aesthetics assessment that predicts the distribution of human ratings. Shu \emph{et al.} (2020) \cite{shu2020} proposed a deep CNN which process aesthetic attributes as privileged information. In another study, Attribute-assisted Multimodal Memory Network (AMM-Net) \cite{li2023b} extracts attributes to model the interactions between visual and textual utilities.

Assessing the aesthetic quality of images involves numerous factors that contribute to preferences, and while many of these factors are difficult to quantify, there are some known aesthetic attributes that influence preferences. Previous studies have investigated the aesthetic value of images in conjunction with these attributes. Recently, deep neural networks based on multi-task learning have been employed to tackle this task, treating it as a multi-task problem that simultaneously predicts an overall aesthetic score and multiple attribute scores. For instance, Kong \emph{et al.} (2016) \cite{kong2016} introduced the AADB dataset, which includes overall aesthetic scores and scores for eleven attributes of photos. In their study, Kong \emph{et al.} (2016) \cite{kong2016} developed a multi-task neural network by fine-tuning AlexNet \cite{krizhevsky2012}, and training a Siamese network \cite{chopra2005} to predict aesthetic ratings. Subsequent studies have also utilized the AADB dataset for further research in this field. For example, Hou \emph{et al.} (2017) \cite{hou2017} applied the squared earth mover's distance-based loss for training, and compared different deep networks including AlexNet, VGG16 \cite{simonyan2015}, and a wide residual network \cite{zagoruyko2016}, and found that fine-tuning a VGG-based model achieved the best performance. Li \emph{et al.} (2019) \cite{li2019} proposed a multi-task model which learns image aesthetics and personality traits. The multi-task network proposed by Celona \emph{et al.} (2022) \cite{celona2022} is able to predict aesthetic score as well as style and composition attributes.   

Pan \emph{et al.} (2019) \cite{pan2019} proposed a neural network architecture based on adversarial learning inspired by generative adversarial networks \cite{goodfellow2014}. This is a multi-task deep CNN namely ``rating network'' which learns the aesthetic score and attributes simultaneously. While the rating network plays the role of ``generator'', a ``discriminator'' tries to distinguish the predictions of multi-task network from the real values. This model outperforms previous approaches and is currently considered as the state-of-the-art method for predicting the overall aesthetic scores on the AADB dataset in multi-task aesthetic prediction.

Since most of the images rated null (neutral) for three attributes (symmetry, repetition and motion blur) in the AADB dataset (Figure~\ref{aadb_attributes}), some studies \cite{malu2017, abdenebaoui2018, reddy2020, li2022} have chosen to exclude these attributes from their multi-task models. For instance, Malu \emph{et al.} (2017) \cite{malu2017} developed a multi-task CNN based on ResNet-50 \cite{he2016} which simultaneously learns the eight aesthetic attributes along with the overall aesthetic score. They also examined the salient regions for the corresponding attribute and applied the gradient based visualization technique \cite{zhou2016}. Abdenebaoui \emph{et al.} (2018) \cite{abdenebaoui2018} used a deep CNN that predicts technical quality, high-level semantic quality, and a detailed description of photographic rules. Reddy \emph{et al.} (2020) \cite{reddy2020} proposed a multi-task network based on EfficientNet \cite{tan2019} for the same purpose, along with a visualization technique and activation maps generated using Gradient-weighted Class Activation Mapping (Grad-CAM) \cite{selvaraju2017} to generate activation maps. Recently, Li \emph{et al.} (2022)\cite{li2022} presented a hierarchical image aesthetic attribute prediction model. Theme-Aware Visual Attribute Reasoning (TAVAR) model, introduced by Li \emph{et al.} (2023) \cite{li2023a}, can predict six attributes of the AADB dataset.   

Besides, Li \emph{et al.} (2020) \cite{li2020} proposed a multi-task deep learning framework that takes into account an individual's personality in modeling their subjective preferences. Liu \emph{et al.} (2020) \cite{liu2020a} developed an aesthetics-based saliency network in a multi-tasking setting. The aesthetic evaluation system proposed by Jiang \emph{et al.} (2021) \cite{jiang2021} outputs the image style label and three forms of aesthetic evaluation results for an image. 

More recently, another image dataset, namely the Explainable Visual Aesthetics (EVA) \cite{kang2020}, has been released, which includes overall aesthetic scores and attribute scores. Although there are a few studies that have used this dataset for aesthetics research, their models only predict overall aesthetic scores \cite{shaham2021, duan2022, li2023, li2023a}. Therefore, our study is the first multi-task neural network that can make predictions on the EVA dataset.   

Current multi-task learning approaches have demonstrated the feasibility of predicting ratings in the AADB dataset by utilizing all the available attributes instead of excluding some. In line with this, we employ all the attributes in the AADB dataset and develop a neural network architecture that is both efficient and effective in predicting overall aesthetic scores of images. Moreover, we evaluate our multi-task CNN on the EVA dataset to further assess its performance. Our multi-task CNN provides predictions for the EVA dataset, serving as a baseline for future research in this area.

\begin{figure*}[t]
%\vspace{.3in}
\centering
\includegraphics[scale=0.29]{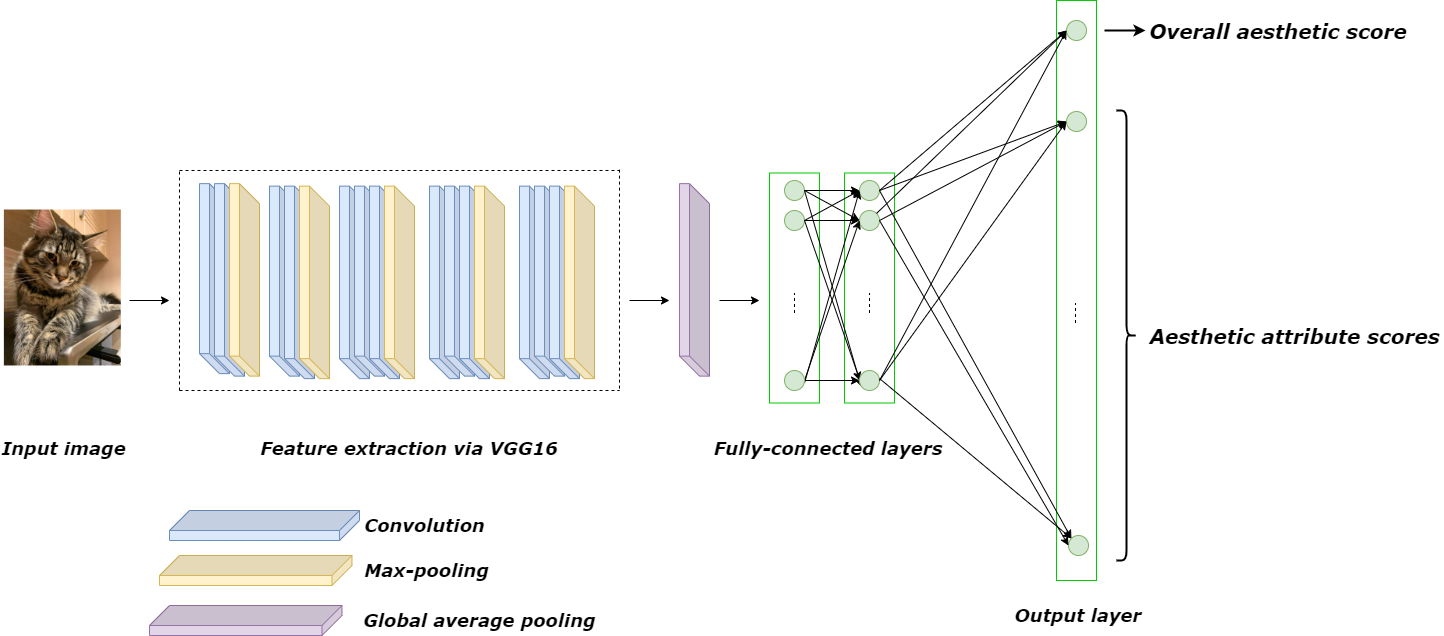}
%\vspace{.3in}
\caption{The general architecture of our multi-task convolutional neural network.}
\label{model}
\end{figure*} 

\section{Proposed Multi-Task Convolutional Neural Network} \label{nn}

We propose a deep multi-task CNN that jointly learns the overall aesthetic score and the aesthetic-related attributes for images during training. This allows the resulting neural network to simultaneously predict multiple scores for an image. We train our deep neural network directly from RGB images, and it is based on the VGG16 pretrained network to extract features. The prior multi-task approaches mentioned in Section \ref{related-work} have already proved that using a pretrained network is a better option than training a neural network from scratch, since both the AADB and EVA datasets do not include large numbers of images. We conducted experiments on several candidate pre-trained CNNs to determine the optimal architecture for our task, and selected the model with the highest performance. It's worth noting that we also tested the Transformer model as a backbone, but observed a tendency for overfitting in our application, likely due to their extensive number of parameters. Consequently, we prioritized a model that not only performs well but also maintains computational efficiency.   

The proposed neural network takes images as input and uses VGG16 to extract feature representations, as shown in Figure~\ref{model}. We removed the fully-connected layers in VGG16 and used the five blocks of convolutional layers. We added a global average pooling layer to the output of the last convolutional block of VGG16. The resulting feature maps are fed into two fully-connected layers with ReLU activation function \cite{nair2010, glorot2011}, consisting of 128 and 64 hidden units, respectively. To prevent overfitting, we applied dropout \cite{srivastava2014} with a rate of 0.35 to the second fully-connected layer with 64 hidden units, which precedes the output layer. The architecture of our neural network consists of multiple units in the output layer, one for predicting the overall aesthetic score and additional units for predicting attribute scores. For the AADB dataset, which has 11 attributes, there are 12 output units in total. On the other hand, the EVA dataset has 4 attributes, so our model includes 5 output units. For more information about the datasets, please refer to Section~\ref{datasets}. The output layer applies sigmoid activation function, and all the output units share the same hidden representation. Notably, we also designed our multi-task CNN with separate output layers: one for the overall aesthetic score and one for each attribute. Interestingly, both architectures perform similarly for predicting the overall aesthetic score. However, we found that the architecture with a single output layer outperforms the one with separate output layers when it comes to predicting the attribute scores. 

\section{Experimental Setup}\label{setup}

\subsection{Datasets}\label{datasets}
Both AADB and EVA datasets we use in this study provide aesthetic attribute scores that are suitable for regression modeling, in addition to the overall aesthetic scores. The Aesthetic Visual Analysis (AVA) dataset \cite{murray2012} is another widely used benchmark in aesthetics research. However, the AVA dataset only provides binary labels for attributes, which is not suitable for our proposed framework as it requires rating scores. Additionally, many images in the AVA dataset are either heavily edited or synthetic, which limits its applicability. In contrast, the AADB dataset provides a more balanced distribution of professional and consumer photos, as well as a more diverse range of photo qualities \cite{kong2016}. Consequently, we utilize the AADB and EVA datasets for training and evaluating our multi-task CNN. These datasets are described below.

\textbf{AADB.}
We utilize the Aesthetics with Attributes Database (AADB) \cite{kong2016}, an image aesthetic benchmark containing 10,000 RGB images of size 256 $\times$ 256 collected from the Flickr website. Each image has overall aesthetic scores provided by 5 different raters. The scores are on a scale of 1 to 5, with 5 being the most aesthetically pleasing score.  Additionally, there are eleven attributes that are known to impact aesthetic judgments according to professional photographers. In this dataset, every image has also scores for each attribute. These attributes are balancing element, interesting content, color harmony, shallow depth of field, good lighting, motion blur, object emphasis, rule of thirds, vivid color, repetition, and symmetry. The raters indicated whether each attribute has a positive, negative, or null (zero) effect on the aesthetics of an image, except for repetition and symmetry where only the presence or absence of the attribute is rated. 

\begin{figure}[h]
%\vspace{.1in}
\centering
\includegraphics[scale=0.5]{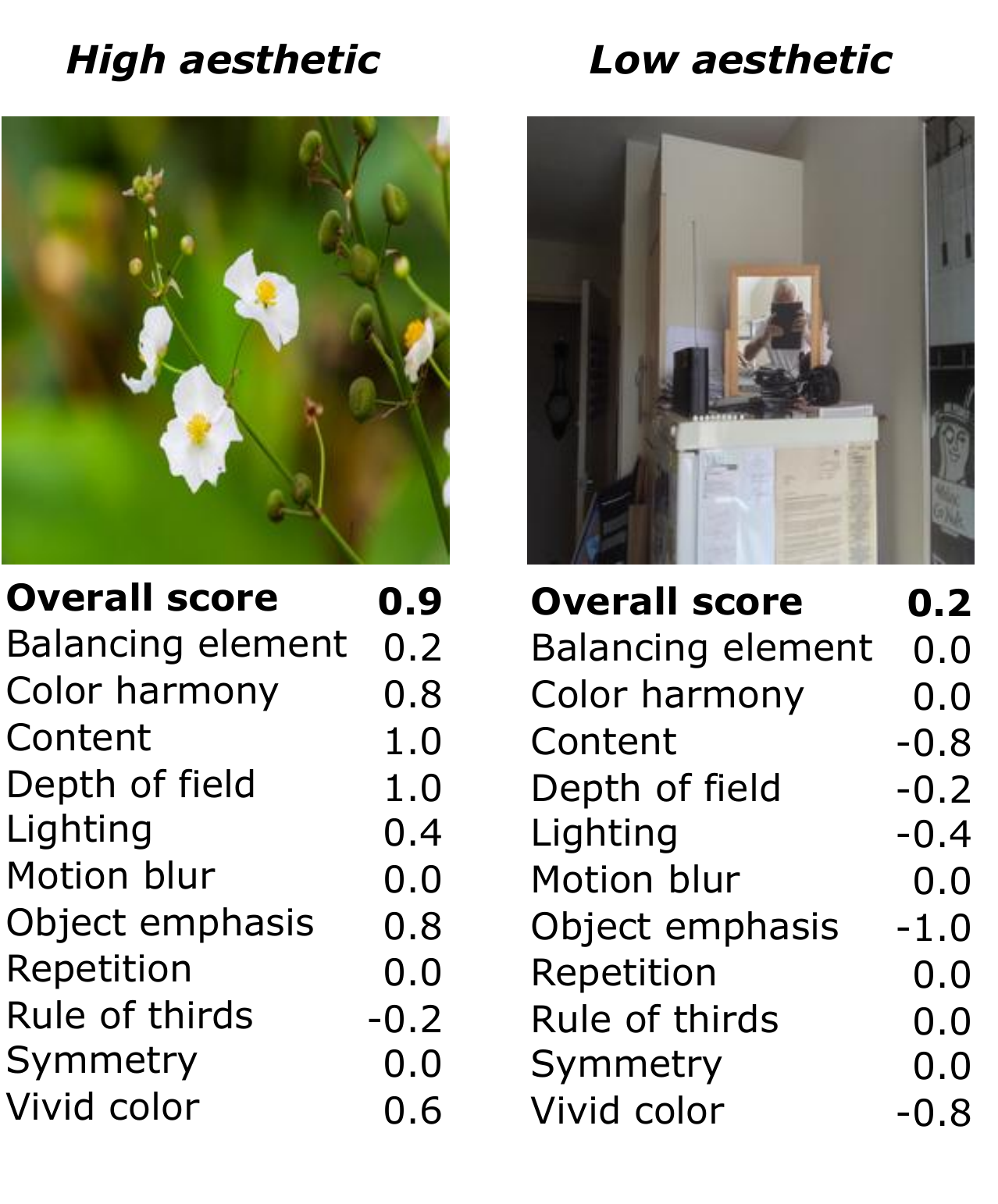}
%\vspace{.1in}
\caption{Example images from the training set of the AADB dataset. Each image has overall aesthetic score and scores for 11 attributes. {\em (Left)} 
High aesthetic: An image rated high on overall aesthetic score. {\em (Right)} Low aesthetic: An image rated low on overall aesthetic score.}
\label{aadb_examples}
\end{figure}

To obtain the ground-truth scores for each image in the AADB dataset, Kong \emph{et al.} (2016)  \cite{kong2016} calculated the average aesthetic scores provided by five different raters. Since only the average scores are reported, the individual rater scores are not available in the dataset. Then, the average scores are normalized to the range of [0,1], while all the attributes except for repetition and symmetry are normalized to the range of [-1,1]. Repetition and symmetry are normalized to the range of [0,1]. Two sample images from the AADB dataset, showcasing examples of both low and high aesthetics, are shown in Figure \ref{aadb_examples}. 

\begin{figure}[h]
%\vspace{.3in}
\centering
\includegraphics[scale=0.29]{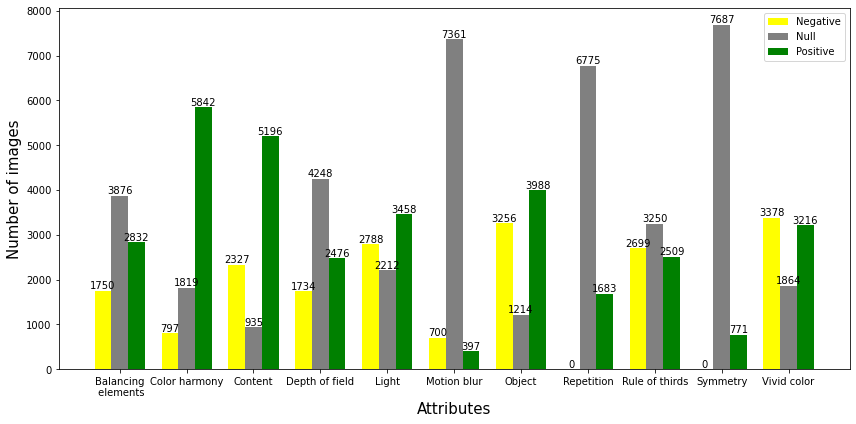}
%\vspace{.3in}
\caption{Visualization of image attribute data in the training set of AADB dataset illustrating the distribution of negative, null, and positive levels for each attribute \cite{kong2016}.}
\label{aadb_attributes}
\end{figure}

\newpage 
The distribution of the attributes is presented in Figure~ \ref{aadb_attributes}. Among them, the motion blur, repetition, and symmetry attributes are mostly rated neutral. Therefore, as mentioned in Section \ref{related-work}, some researchers excluded these three attributes from their multi-task neural networks. However, the motion blur attribute has both negative (700) and positive (397) scores, which may still provide useful information. Similarly, the repetition attribute has 1683 positive scores, while the symmetry attribute has 771 positive scores. It is worth noting that raters were not allowed to give negative scores for repetition and symmetry.  

The AADB dataset has been split into three subsets: 500 images for validation, 1000 images for testing, and the remaining images for training, following the official partition  \cite{kong2016}. For our experiments, we use this partition to train and test our multi-task CNN, allowing for direct comparison with other approaches.

\begin{figure}[h]
%\vspace{.3in}
\centering
\includegraphics[scale=0.49]{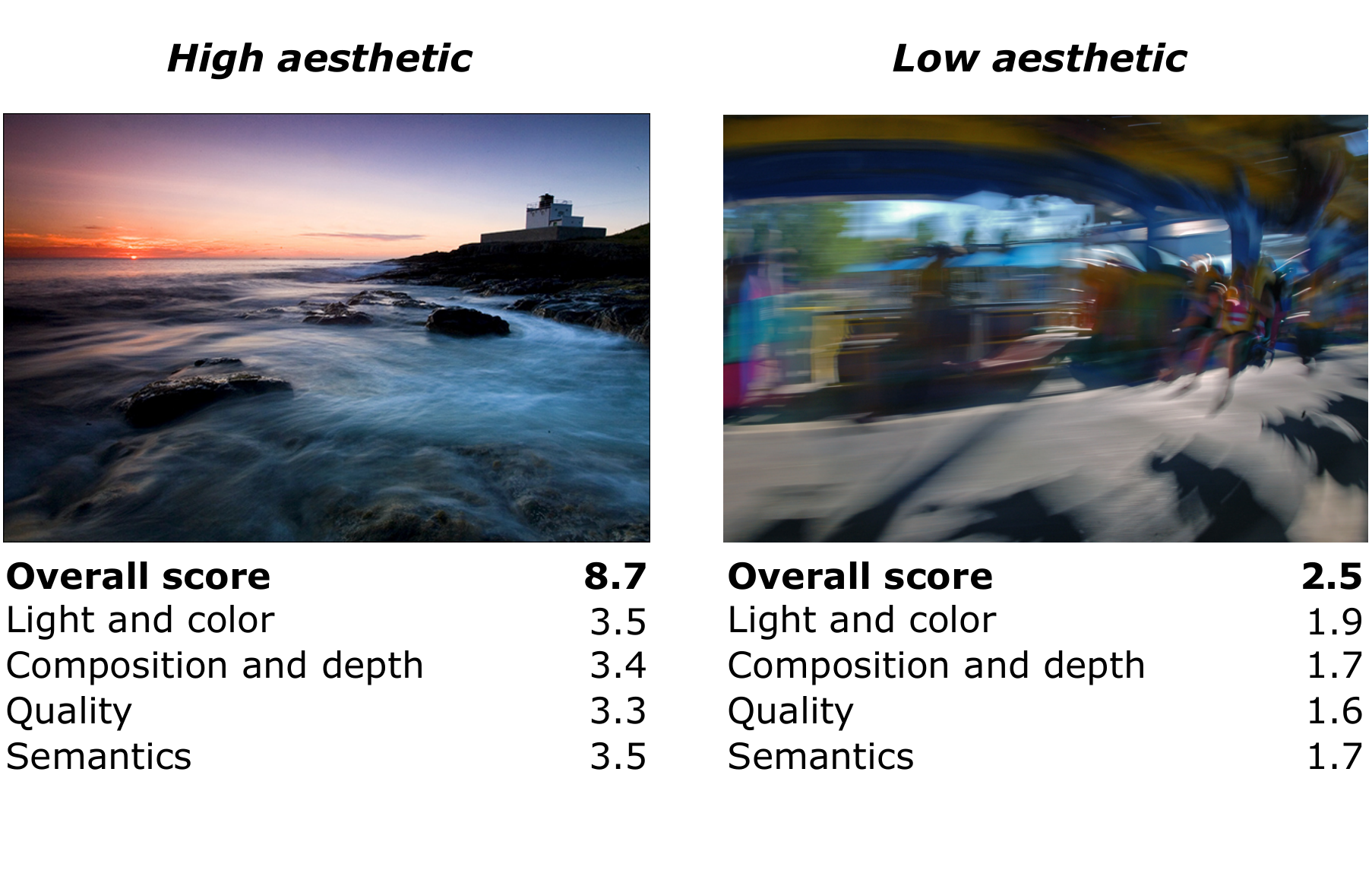}
%\vspace{.3in}
\caption{Example images from the training set of the EVA dataset. Each image has overall aesthetic score and scores for 4 attributes. {\em (Left)} High aesthetic: 
An image rated high on overall aesthetic score. {\em (Right)} Low aesthetic: An image rated low on overall aesthetic score.}
\label{eva_examples}
\end{figure}

\textbf{EVA.} The Explainable Visual Aesthetics (EVA) dataset \cite{kang2020} contains 4070 images, each rated by at least 30 participants. The EVA dataset overcomes the limitations of previous datasets by including images with 30 to 40 votes per image, collected using a disciplined approach to avoid noisy labels due to misinterpretations of the tasks or limited number of votes per image \cite{kang2020}. Each image has an aesthetic quality rating with an 11-point discrete scale. The extremes of the scale are labelled as "least beautiful" (corresponding to 0) and "most beautiful" (corresponding to 10). The EVA dataset contains four attributes: light and color, composition and depth, quality, and semantics of the image. For each attribute, the images were rated on a four-level Likert scale (very bad, bad, good, and very good). Two sample images from the EVA dataset, showcasing examples of both low and high aesthetics, are shown in Figure~ \ref{eva_examples}.  

In contrast to the AADB dataset, Kang \emph{et al.} (2020) \cite{kang2020} reported all ratings from the participants. So, we calculated the average scores for each image. Unlike the AADB dataset, which has predetermined train-validation-test splits, there is no official train-validation-test split for the EVA dataset, since Kang \emph{et al.} (2020) did not use any neural network. However, studies focusing on predicting only the overall aesthetic scores (see Section \ref{related-work}) on the EVA dataset have utilized different training and testing splits. For example, Duan \emph{et al.} (2022) \cite{duan2022} and Li \emph{et al.} (2023a) \cite{li2023a} employed a split of 3,500 training images and 570 testing images, while Li \emph{et al.} (2023) \cite{li2023} used 4,500 training images and 601 testing images. Similarly, Shaham \emph{et al.} (2021)  \cite{shaham2021} utilized a split of 2,940 training images and 611 testing images.

\subsection{Implementation Details}\label{implementation}
We initialize the fully-connected layer weights in our multi-task CNN with the Glorot uniform initializer \cite{glorot2010}. We use \textit{mean squared error} as the loss function on the training set $X$ to minimize the error between the predictions and the ground-truth values: 

\begin{equation}
E(W|X) = \frac{1}{n}\sum_{i=1}^{n}(y_i - \hat{y_i})^2
\end{equation}
where $n$ is the number of samples in the training set, $y_i$ are the ground-truth scores and $\hat{y_i}$ are the predictions generated by Eq. \ref{function}. In our multi-task model, we examined the implementation of a weighted loss function. However, our empirical evaluations indicated no significant improvement in model performance. Consequently, we do not implement a weighted loss function in our proposed model. 

Since both datasets we use in this study do not include large numbers of images, we apply horizontal flip as data augmentation. We train our multi-task CNN in two stages. In the first stage, we apply the Adam algorithm \cite{kingma2014} with an initial learning rate of 0.001 and decay constants of 0.9 and 0.999. The VGG16 pretrained network is composed of five blocks, each of which includes convolutional and pooling layers. During the first stage, we freeze the weights for all five blocks and train the multi-task CNN for 5 epochs, with a minibatch size of 64. We closely monitor the training and validation loss during this stage and observe that the model is prone to overfitting if we train it for longer, without any notable improvement in its performance. 

In the second stage, we fine-tune the multi-task CNN by unfreezing the last two convolutional layer in the fourth block of VGG16. We apply the Adam algorithm again, but we adjust its learning rate by using the exponential decay learning rate schedule. The initial learning rate is 0.0001, and it decays every 125 steps with a base of 0.50. We fine-tune the model with this setting for 3 epochs with a minibatch size of 64. 

\section{Results and Discussion}
\label{results}
In this section, we provide a comprehensive evaluation of our proposed multi-task CNN on both datasets, with an emphasis on its efficiency in assessing image aesthetics. We explore the impact of fine-tuning and analyze the results for both overall and attribute scores predicted by our model. Additionally, we compare the performance of our multi-task approach with a single-task setting for predicting the overall aesthetic scores of images using the same neural network architecture.  

\subsection{Performance analysis on the AADB dataset}

\subsubsection{Model evaluation and comparison with the state-of-the-art}

Table~\ref{comparison} provides an overview of the performances achieved by the studies in the literature which use the AADB dataset to develop multi-task deep neural networks. These neural networks learn the eleven attributes of the AADB dataset along with the overall aesthetic score of images. 

\begin{table}[h]
\caption{Comparison of performances achieved by previous multi-task neural networks and our proposed multi-task CNN on the test set of the AADB dataset.}
\label{comparison}
\centering
\begin{tabular}{ll}
\textbf{METHODS} & $\boldsymbol{\rho}$\\
\hline
(Kong \emph{et al.}, 2016) & 0.6782 \\
(Hou \emph{et al.}, 2017) & 0.6889 \\
(Pan \emph{et al.}, 2019)$^a$ & 0.6927 \\
(Pan \emph{et al.}, 2019)$^b$ & 0.7041 \\
Ours & \textbf{0.7067}\\ 
\end{tabular}

\footnotesize{$^a$ The multi-task neural network, $^b$ The multi-task neural network with the adversarial learning setting.}
\end{table}

%\begin{table*}[h]
%\caption{The Number of Parameters for the Multi-Task Neural Network Proposed by Pan \emph{et al.} (2019) vs. Our Multi-Task Neural Network.}
%\label{complexity-comparison}
%\centering
%\begin{tabular}{|l|c|c|c|c|}
%\hline
%& \multicolumn{2}{c|} {\textbf{Pan \emph{et al. (2019)}}} & \multicolumn{2}{c|} {\textbf{Our Method}} \\
%\cline{2-5}
%& \textbf{Architecture} & \textbf{Parameters} & \textbf{Architecture} & \textbf{Parameters} \\
%\hline
%Pre-trained Network & ResNet-50 & 23,587,712 & VGG16 & \textbf{14,714,688} \\
%\hline
%Fully-Connected Layers$^*$ & (512-128) & 1,114,752 & (128-64) & \textbf{73,920} \\
%\hline
%Output Layer$^\dagger$ & (12) & 1548 & (12) & \textbf{780} \\
%\hline
%\end{tabular}

%\footnotesize{$^*$ The number of hidden units for two fully-connected layers, respectively.}

%\footnotesize{$^\dagger$ The number of output units.}
%\end{table*}

\begin{table*}[h]
\caption{The Number of Parameters for the Multi-Task Neural Network Proposed by Pan \emph{et al.} (2019) vs. Our Multi-Task Neural Network.}
\label{complexity-comparison}
\centering
\begin{tabular}{lcccc}
& \multicolumn{2}{c} {\textbf{Pan \emph{et al. (2019)}}} & \multicolumn{2}{c} {\textbf{Our Method}} \\
\cline{2-5}
& \textbf{Architecture} & \textbf{Parameters} & \textbf{Architecture} & \textbf{Parameters} \\
Pre-trained Network & ResNet-50 & 23,587,712 & VGG16 & \textbf{14,714,688} \\
Fully-Connected Layers$^*$ & (512-128) & 1,114,752 & (128-64) & \textbf{73,920} \\
Output Layer$^\dagger$ & (12) & 1548 & (12) & \textbf{780} \\
\end{tabular}

\raggedright
\footnotesize{$^*$ The number of hidden units for two fully-connected layers, respectively.}

\footnotesize{$^\dagger$ The number of output units.}
\end{table*}

To make a comparison between the previous studies and our multi-task CNN, we use Spearman's rank correlation coefficient ($\rho$), which is a commonly used metric in this field. Table~\ref{comparison} summarizes the $\rho$ values reported in each study, which represent the correlation between the estimated overall aesthetic scores by the multi-task neural network and the corresponding ground-truth scores in the test set. We calculate this correlation using the overall aesthetic scores predicted by our multi-task CNN and find it to be significant at p $<$ 0.01. This allows us to compare the performance of our model to those in the literature. 

As shown in Table~\ref{comparison}, there has been a slight improvement in the correlation between predicted overall scores and ground-truth scores over the years.  The approach proposed by Pan \emph{et al.} (2019) \cite{pan2019} has resulted in the highest correlation achieved thus far. Their study includes two methods, the first of which is a multi-task deep neural network that achieves a Spearman's rank correlation of 0.6927. The second one takes the first method one step forward by updating it with an adversarial setting, as described in Section \ref{related-work}. Compared to these methods, our multi-task CNN outperforms the first method (0.7067 $>$ 0.6927). When we compare our model with the adversarial learning setting proposed by Pan \emph{et al.} (2019) \cite{pan2019}, we find that our neural network outperforms theirs again (0.7067 $>$ 0.7041). 

In addition to achieving the highest Spearman's rank correlation, our multi-task CNN has other advantages over the state-of-the-art approach proposed by Pan et al. (2019) \cite{pan2019}.  Table~\ref{complexity-comparison} compares the neural network architectures of current state-of-the-art \cite{pan2019} and our multi-task CNN by taking into account the number of parameters. While Pan \emph{et al.} (2019) \cite{pan2019} uses ResNet-50 for feature extraction, we utilize VGG16, which has fewer parameters for this particular problem. Furthermore, the fully-connected layers of our model have significantly fewer parameters compared to those in Pan \emph{et al.} (2019) \cite{pan2019}. Specifically, our model's fully-connected layers have around 15 times fewer parameters in the neural network compared to their model. Similarly, the output layer of our multi-task CNN has fewer parameters, too. Moreover, the adversarial setting used in Pan \emph{et al.} (2019) \cite{pan2019} makes the training of their neural network more complex.

\begin{figure}[h]
%\vspace{.3in}
\centering
\includegraphics[scale=0.5]{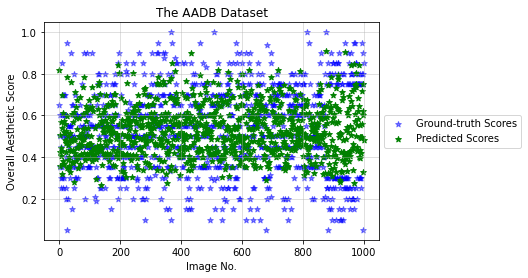}
%\vspace{.3in}
\caption{Visualization of model predictions: A scatter plot comparing the actual overall aesthetic scores of test images in the AADB dataset to the predicted scores generated by our multi-task CNN.}
\label{aadb_prediction_graph}
\end{figure}

We also evaluate the predictions made by our multi-task CNN and investigate the issue of overfitting. Figure~\ref{aadb_prediction_graph} shows that our model can predict overall aesthetic scores across a wide range. While the ground-truth overall aesthetic scores in the test data range from 0.05 to 1.0, our model's predictions range from 0.26 to 0.90. We also report the frequencies and percentages of ground-truth overall aesthetic scores in Table~\ref{aadb_frequencies} for different intervals in the test data. Based on these data, our model's predictions are not very good for 39 samples falling in intervals [0.05-0.10] and [0.10-0.20], and for 15 samples falling in the interval [0.90-1.00]. In other words, our multi-task CNN can make successful predictions for approximately 95$\%$ of test data. This indicates that our multi-task CNN is able to make predictions for the majority of the test data, with only a small percentage of samples falling outside of its prediction range. These results also indicate that there is no issue of overfitting.

\begin{figure*}[h]
%\vspace{.3in}
\centering
\includegraphics[scale=0.64]{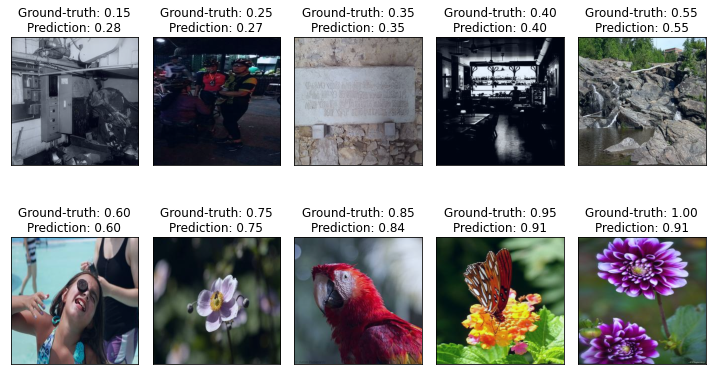}
%\vspace{.3in}
\caption{Comparison of ground-truth overall aesthetic scores and corresponding predictions by our multi-task CNN on the test data of the AADB dataset. This figure shows \underline{the most successful} predictions ranging from low aesthetic images to high aesthetic images.}
\label{aadb_results_best}
\end{figure*} 

\begin{figure*}[h!]
%\vspace{.3in}
\centering
\includegraphics[scale=0.64]{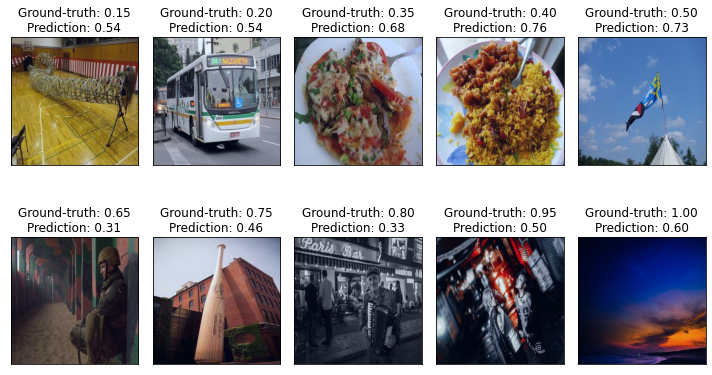}
%\vspace{.3in}
\caption{Comparison of ground-truth overall aesthetic scores and corresponding predictions by our multi-task CNN on the test data of the AADB dataset. This figure shows \underline{the least successful} predictions ranging from low aesthetic images to high aesthetic images.}
\label{aadb_results_worst}
\end{figure*} 

In order to further evaluate the performance of our multi-task CNN, we visually examine its predictions and present the most successful predictions in Figure~\ref{aadb_results_best} and the least successful ones in Figure~\ref{aadb_results_worst}. As shown in Figure~\ref{aadb_results_best}, our multi-task CNN exhibits remarkable predictive performance, with the exception of one low-aesthetic image with the ground-truth score of 0.15. On the other hand, in the least successful predictions shown in Figure~\ref{aadb_results_worst}, our model tends to predict high scores to low aesthetic images, while giving lower scores to the high aesthetic images. 

Overall, our multi-task CNN achieves the highest Spearman's rank correlation for overall aesthetic scores while simultaneously predicting scores for 11 attributes. Notably, our approach accomplishes this with fewer parameters, making it more computationally efficient than the state-of-the-art method proposed by Pan \emph{et al.} (2019) \cite{pan2019}. By combining a simplified neural network architecture with superior predictive performance, our approach represents a significant advancement in the field of image aesthetic assessment.

\begin{table}[h]
\caption{The summary of the frequencies and percentages of ground-truth overall aesthetic scores falling into specific intervals within the test data of the AADB dataset.} \label{aadb_frequencies}
\begin{center}
\begin{tabular}{lll}
\textbf{Interval} &\textbf{Frequency} &\textbf{Percentage} \\
\hline \\
0.05 - 0.10 & 4   & 0.004 \\ 
0.10 - 0.20 & 35  & 0.035 \\
0.20 - 0.30 & 79  & 0.079 \\  
0.30 - 0.40 & 147 & 0.147 \\
0.40 - 0.50 & 152 & 0.152 \\
0.50 - 0.60 & 165 & 0.165 \\
0.60 - 0.70 & 224 & 0.224 \\
0.70 - 0.80 & 76  & 0.076 \\
0.80 - 0.90 & 103 & 0.103 \\
0.90 - 1.00 & 15  & 0.015 \\
\end{tabular}
\end{center}
\end{table}

\subsubsection{Comparison with human performance}
In addition to evaluating the performance of our multi-task CNN in rating image aesthetics, we compare its results with human performance on the AADB dataset. Kong \emph{et al.} (2016) \cite{kong2016} previously reported the Spearman's rank correlation between each individual's ratings and the ground-truth average score on this dataset. A subset of raters was selected based on the number of images they have rated. In their study, they found that the more images an individual rated, the more stable their aesthetic score rankings became. We utilize this data and compare it to the performance of our model, as shown in Table~\ref{human}.

\begin{table}[h]
\caption{The Comparison Between Human Performance and Our Multi-Task CNN on the AADB Database.} \label{human}
\begin{center}
\begin{tabular}{lll}
\textbf{\#} \textbf{images rated} & \textbf{\#} \textbf{raters} & \bm{$\rho$} \\
\hline \\
(0,100)               &190 &0.6738 \\ 
${[}100,200)$             &65  &0.7013\\
${[}200, \infty)$             &42  &0.7112\\
Our method       &-   & \textbf{0.7067}\\
\end{tabular}
\end{center}
\end{table}

Based on these correlations, we see that when the number of images rated by the same observer increases, human performance becomes better. On the other hand, it is also clear that our multi-task CNN performs above the level of human consistency averaged across all raters. Only when compared to the more experienced raters (i.e., the 42 raters who rated $>$200 images), our model performs slightly less. Our experiments demonstrate that our multi-task CNN achieves near-human performance in predicting the overall aesthetic scores on the AADB dataset. This narrows the performance gap between machines and humans in this domain.

\subsubsection{Attribute predictions and the fine-tuning effect}
\begin{figure*}[h]
\vspace{.3in}
\centering
\includegraphics[scale=0.4]{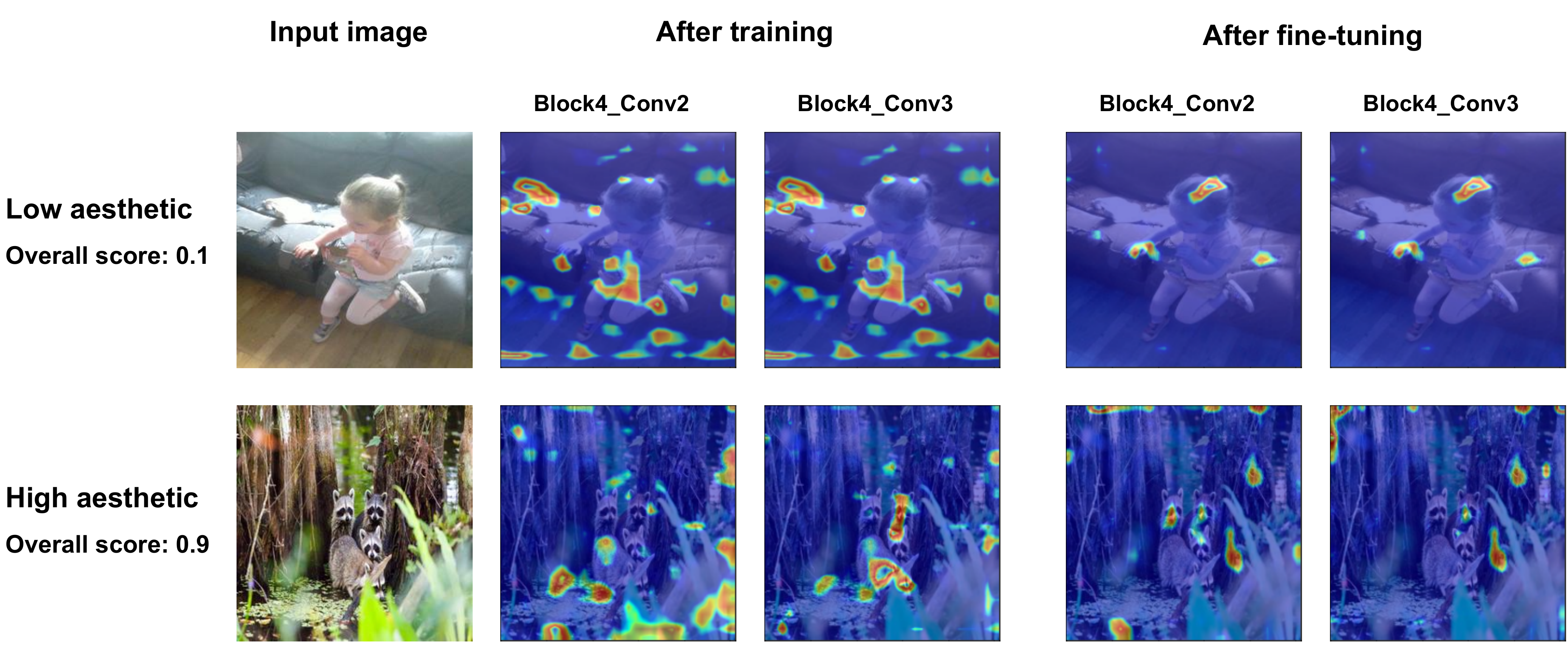}
\vspace{.3in}
\caption{The activation maps for two input images from the test set of AADB dataset. These maps highlight the regions of the input image that contributed the most to the neural network's prediction. The heatmap is overlaid on top of the input image to provide a visualization of which areas of the image are most relevant for the task.}
\label{activation_maps_aadb}
\end{figure*}

Table~\ref{attributes} displays the Spearman's rank correlations between the ground-truth scores and the corresponding predictions made by our multi-task CNN for each attribute. Moreover, we investigate the effect of fine-tuning and include those results in Table~\ref{attributes}. As described in Section~\ref{implementation}, we train our multi-task CNN, and then we fine-tune the model by unfreezing the last two convolutional layer in the fourth block of VGG16 ($block4\_ conv2$ and $block4\_conv3$). After fine-tuning our model, we observe an increase in the correlations for all attributes except symmetry. We also observe an increase in the correlation for the overall aesthetic score. Moreover, we aim to gain insight into the two convolutional layers that we fine-tune. To this end, we illustrate the activation maps generated using Grad-CAM \cite{selvaraju2017} in Figure~\ref{activation_maps_aadb} for two images from the test set of AADB dataset, one with a low aesthetic score and the other with a high aesthetic score. 

\begin{table}[h]
\caption{Spearman's rank correlations between the ground-truth scores for each attribute and the predictions by our multi-task CNN in the test data of the AADB Dataset. This table shows the correlations after training and after fine-tuning separately, in addition to the correlations for the overall aesthetic score. } \label{attributes}
\begin{center}
\begin{tabular}{lll}
\textbf{Attributes}  & \textbf{Training} &\textbf{Fine-tuning}  \\
\hline \\
Content            & 0.546 & 0.593  \\ 
Vivid Color        & 0.617 & 0.669  \\
Object Emphasis    & 0.600 & 0.639  \\
Color Harmony      & 0.437 & 0.484   \\
Depth of Field     & 0.466 & 0.497   \\
Lighting           & 0.396 & 0.445   \\
Balancing Elements & 0.264 & 0.267   \\
Rule of Thirds     & 0.216 & 0.235   \\
Motion Blur        & 0.098 & 0.109   \\
Symmetry           & 0.194 & 0.177   \\
Repetition         & 0.322 & 0.355   \\
\textbf{Overall}   & \textbf{0.650} & \textbf{0.707} \\
\end{tabular}
\end{center}
\end{table}

\begin{figure*}[h]
%\vspace{.3in}
\centering
\includegraphics[scale=0.70]{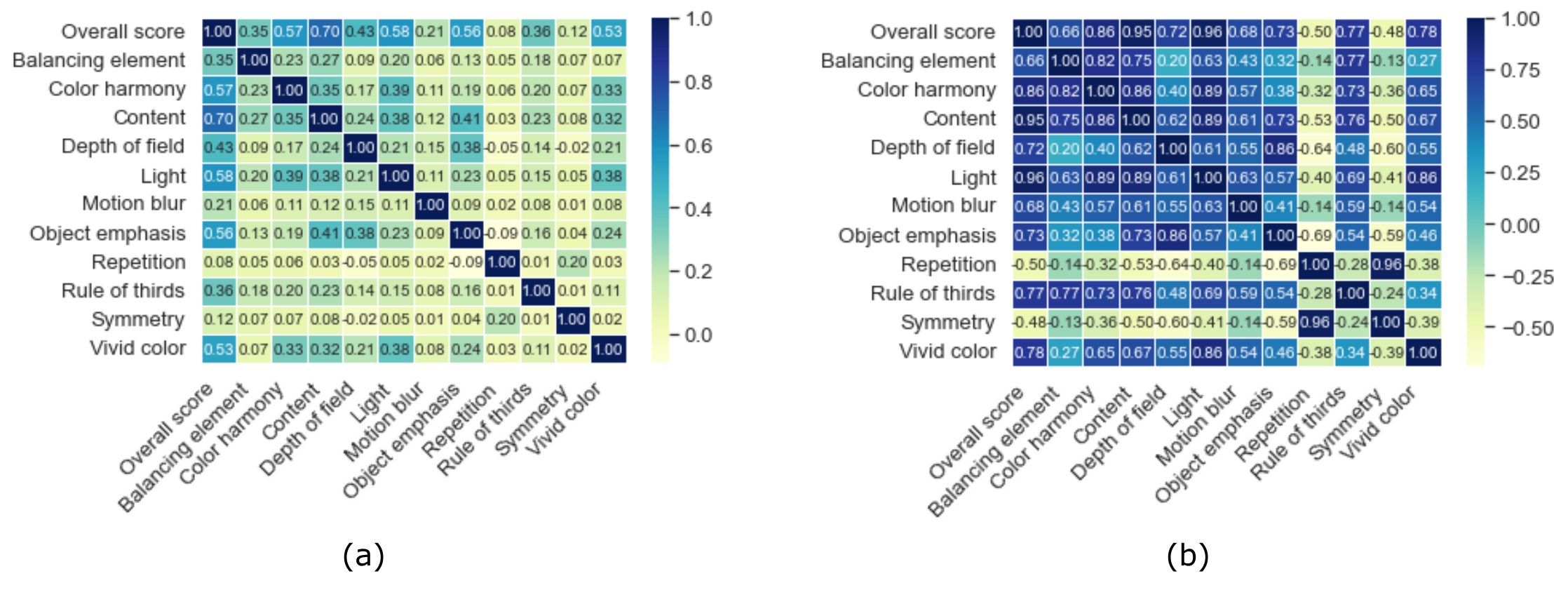}
%\vspace{.3in}
\caption{Spearman's rank correlations between the overall aesthetic scores and the attribute scores on the AADB dataset. \emph{(a)}: The ground-truth scores on all dataset,  \emph{(b)}: The predictions by our multi-task CNN on the test data.}
\label{correlations_aadb}
\end{figure*} 

We report the Spearman's rank correlations on the AADB dataset in Figure~\ref{correlations_aadb}. This figure shows the correlations for the ground-truth scores of all dataset on the left side, whereas on the right side, we present the correlations for the predictions on the test data made by our multi-task CNN. Our model's highest correlations among all attributes are for the light ($\rho$=0.96) and content ($\rho$=0.95) attributes. This finding is consistent with the AADB dataset, where content has the highest correlation with overall aesthetic scores ($\rho$=0.70) and light follows in second place ($\rho$=0.58). Furthermore, when we compare the top-five correlations for our multi-task CNN (light, content, color harmony, vivid color, and rule of thirds attributes), we see the similar results in the AADB dataset indicating our model can capture the relationships between the overall aesthetic scores and the attributes. On the other hand, when we examine the lowest correlations in Figure~\ref{correlations_aadb}, we find that our model also exhibites lower correlations for the motion blur, symmetry, and repetition attributes, consistent with human data. Accordingly, we can conclude that the predictions made by our multi-task CNN closely match human interpretation.

\subsubsection{Single-task versus multi-task setting}\label{versus}
We wondered what would happen if our proposed model were a single-task neural network instead of a multi-task one. In this case, the neural network just learns the overall aesthetic score, not the scores for attributes. We report our result in Table~\ref{single} and compare it to those of Pan \emph{et al.} (2019) \cite{pan2019}. In terms of single-task networks, the Spearman's rank correlation of our method is slightly higher than Pan \emph{et al.} on the test set of AADB dataset. This indicates that our neural network performs slightly better in the single-task setting while utilizing fewer parameters, highlighting the effectiveness of our approach. Furthermore, we also add the multi-task setting results for both model to make a comparison with the single-task one. Both models show that multi-task learning improves the neural network performance, as the Spearman's rank correlations between the predicted aesthetic scores and ground-truth overall aesthetic scores are consistently higher for the multi-task neural networks than for the single-task ones.    

\begin{table}[h]
\caption{The performance comparison between single-task and multi-task neural networks in terms of Spearman's rank correlations.} \label{single}
\begin{center}
\begin{tabular}{lll}
\textbf{METHODS} &\textbf{Single-task} &\textbf{Multi-task} \\
\hline \\
(Pan \emph{et al.}, 2019)&0.6833 &0.6927 \\ 
Ours  &\textbf{0.6890} &\textbf{0.7067}\\
\end{tabular}
\end{center}
\end{table}

\subsection{Performance analysis on the EVA dataset}

\subsubsection{Model evaluation}
The second benckmark we use in this study, the EVA dataset, provides access to all participant's ratings. To investigate the performance of our multi-task CNN on this dataset, first, we calculated the average score for each image with respect to each attribute and overall aesthetic score. Table~\ref{eva_averages} reports the minimum and maximum averages for each attribute and overall aesthetic score in the EVA dataset.  

\begin{table}[h]
\caption{The summary of the averages for the ground-truth overall aesthetic scores and attribute scores in the EVA dataset.} \label{eva_averages}
\begin{center}
\begin{tabular}{lll}
\textbf{Score} &\textbf{Minimum} &\textbf{Maximum} \\
\hline \\
Overall               & 1.764 & 9.032 \\ 
Light and color       & 1.636 & 3.657 \\
Composition and depth & 1.558 & 3.656 \\  
Quality               & 1.441 & 3.593 \\
Semantics             & 1.735 & 3.645 \\
\end{tabular}
\end{center}
\end{table}

Since there are four attributes in the EVA dataset (light and color, composition and depth, quality, semantics), we modify the output layer of our multi-task CNN to include five units (one for the overall aesthetic score and one for each attribute). Consequently, the output layer of our multi-task CNN consists of 325 parameters for the EVA dataset. We applied dropout \cite{srivastava2014} with a rate of 0.25 to the second fully-connected layer with 64 hidden units, which precedes the output layer. Also, since there is no official train-test split for the EVA dataset, we follow the two studies \cite{duan2022,li2023a} which use 3,500 images for training and 570 for testing. Table~\ref{EVA_single} presents the performance of our multi-task CNN on this dataset, also highlighting the effect of fine-tuning. This table summarizes the Spearman's rank correlations between the estimated overall aesthetic scores by our multi-task CNN and the corresponding ground-truth scores in the test set. We also evaluated the model's performance in the single-task setting and observed that the multi-task setting outperforms it. Consistent with the findings in Section~\ref{versus}, we note that predicting the attributes along with the overall aesthetic score has a positive effect on the overall score for the same neural network architecture.

\begin{table}[h]
\caption{The performance of our multi-task CNN on the test set of the EVA dataset and comparison with the single-task setting in terms of Spearman's rank correlations. The table presents the results obtained after training and after fine-tuning.}\label{EVA_single}
\begin{center}
\begin{tabular}{lll}
 & \textbf{Training} & \textbf{Fine-tuning} \\
\hline \\
Multi-task  &0.600 & \textbf{0.695} \\ 
Single-task &0.604 & 0.675          \\
\end{tabular}
\end{center}
\end{table}

\begin{figure}[h]
%\vspace{.3in}
\centering
\includegraphics[scale=0.5]{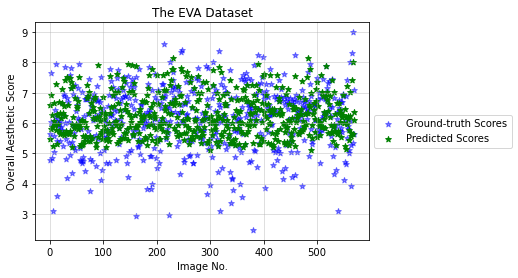}
%\vspace{.3in}
\caption{Visualization of model predictions: A scatter plot comparing the actual overall aesthetic scores of test images on the EVA dataset to the predicted scores generated by our multi-task CNN.}
\label{eva_prediction_graph}
\end{figure}

Similar to the evaluation of the AADB dataset in the previous section, we evaluate the predictions made by our multi-task CNN and investigate the issue of overfitting. We compare the actual overall aesthetic scores of test images in the EVA dataset to the predicted scores generated by our model in Figure~\ref{eva_prediction_graph}. While the ground-truth overall aesthetic scores in the test data range from 2.46 to 9.0, our model's predictions range from 5.09 to 8.13. We also report the frequencies and percentages of ground-truth overall aesthetic scores in Table~\ref{eva_frequencies} for different intervals in the test data. Based on these data, our model's predictions are not very good for 83 samples falling in interval range [1.70-5.00], and for some samples falling in the maximum intervals. In other words, our multi-task CNN can make successful predictions for approximately 85$\%$ of test data. This indicates that our multi-task CNN is able to make predictions for a majority of the test data, with only a small percentage of samples falling outside of its prediction range. These results also indicate that there is no issue of overfitting.

\begin{table}[h]
\caption{The summary of the frequencies and percentages of ground-truth overall aesthetic scores falling into specific intervals within the test data of the EVA dataset.} \label{eva_frequencies}
\begin{center}
\begin{tabular}{lll}
\textbf{Interval} &\textbf{Frequency} &\textbf{Percentage} \\
\hline \\
1.70 - 2.00 & 0   & 0 \\ 
2.00 - 3.00 & 3   & 0.005 \\
3.00 - 4.00 & 16  & 0.028 \\  
4.00 - 5.00 & 64  & 0.112 \\
5.00 - 6.00 & 144 & 0.252 \\
6.00 - 7.00 & 231 & 0.405 \\
7.00 - 8.00 & 100 & 0.175 \\
8.00 - 9.00 & 11  & 0.019 \\
9.00 - 9.50 & 1   & 0.001 \\
\end{tabular}
\end{center}
\end{table}

\begin{figure*}[h]
%\vspace{.3in}
\centering
\includegraphics[scale=0.51]{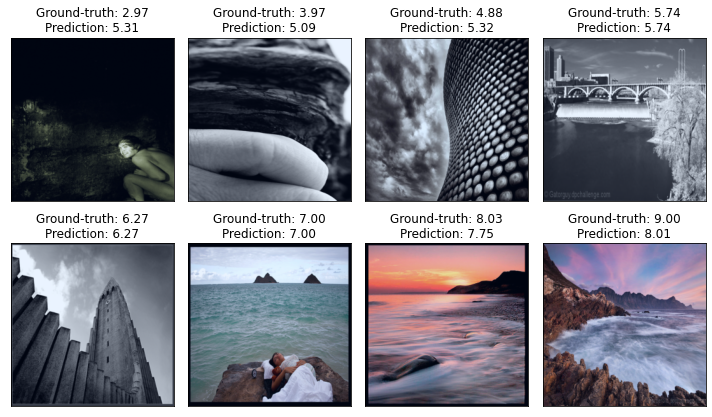}
%\vspace{.3in}
\caption{Comparison of ground-truth overall aesthetic scores and corresponding predictions by our multi-task CNN on the test data of the EVA dataset. This figure shows \underline{the most successful} predictions ranging from low aesthetic images to high aesthetic images.}
\label{eva_results}
\end{figure*} 

\begin{figure*}[h!]
%\vspace{.3in}
\centering
\includegraphics[scale=0.51]{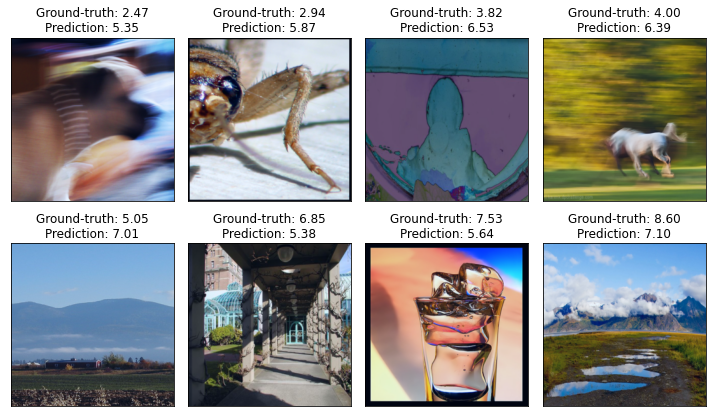}
%\vspace{.3in}
\caption{Comparison of ground-truth overall aesthetic scores and corresponding predictions by our multi-task CNN on the test data of the EVA dataset. This figure shows \underline{the least successful} predictions ranging from low aesthetic images to high aesthetic images.}
\label{eva_results_worst}
\end{figure*} 

We visually examine our model's predictions and
present the most successful predictions in Figure~\ref{eva_results} and the least successful ones in Figure~\ref{eva_results_worst}. As shown in Figure~\ref{eva_results}, our multi-task CNN exhibits remarkable predictive performance for images in the test set of EVA dataset. However, it has more difficulty in predicting scores for the low aesthetic images compared to the AADB dataset. On the other hand, in the least successful predictions shown in Figure~\ref{eva_results_worst}, our model tends to predict high scores for the low aesthetic images, while giving lower scores to the high aesthetic ones. This behavior of the model is consistent with the results obtained from the AADB dataset.

\subsubsection{Attribute predictions and the fine-tuning effect}

\begin{figure*}[h]
%\vspace{.3in}
\centering
\includegraphics[scale=0.3]{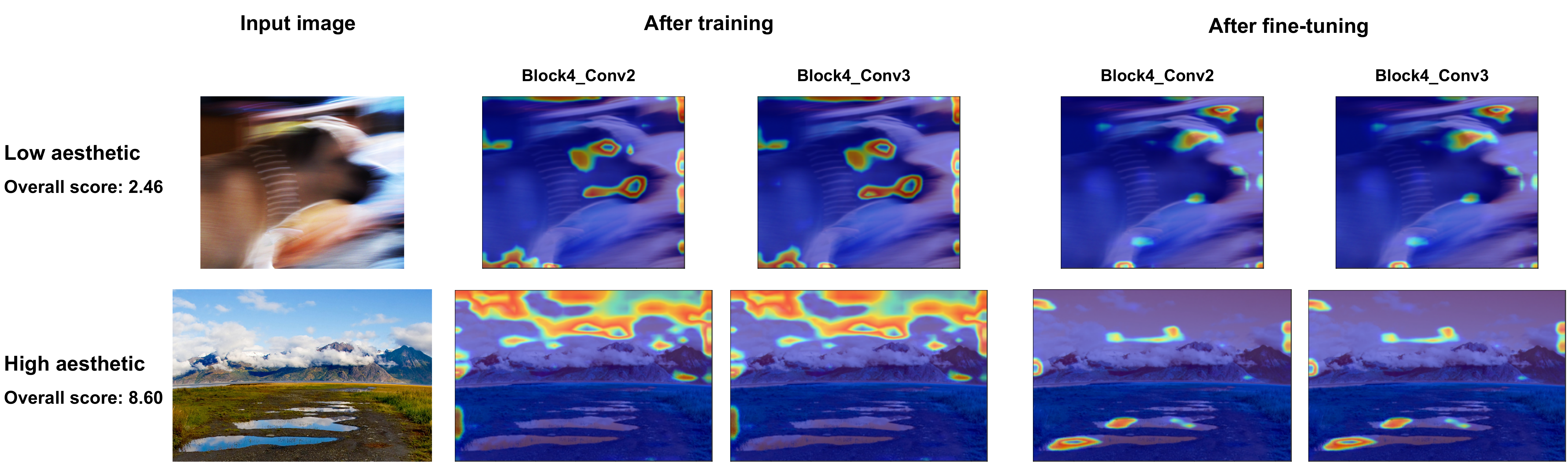}
%\vspace{.3in}
\caption{The activation maps for two input images from the test set of EVA dataset. These maps highlight the regions of the input image that contributed the most to the neural network's prediction. The heatmap is overlaid on top of the input image to provide a visualization of which areas of the image are most relevant for the task.}
\label{activation_maps_eva}
\end{figure*}

Table~\ref{eva_attributes} displays the Spearman's rank correlations between the ground-truth scores and the corresponding predictions made by our multi-task CNN for each attribute. Moreover, we investigate the effect of fine-tuning and include those results in Table~\ref{eva_attributes}. This time, for all the attributes, the correlations increase after fine-tuning. Similar to the evaluation in the AADB dataset, we illustrate the activation maps generated using Grad-CAM \cite{selvaraju2017} in Figure~\ref{activation_maps_eva} for two images from the test set of AADB dataset, one with a low aesthetic score and the other with a high aesthetic score.

\begin{table}[h]
\caption{Spearman's rank correlations between the ground-truth scores for each attribute and the predictions by our multi-task neural network on the test set of the EVA dataset. The table presents the results obtained after training and after fine-tuning.} \label{eva_attributes}
\begin{center}
\begin{tabular}{lll}
\textbf{Attributes}  & \textbf{Training} &\textbf{Fine-tuning}  \\
\hline \\
Light and color       & 0.610 & 0.709  \\ 
Composition and depth & 0.571 & 0.655  \\
Quality               & 0.463 & 0.548  \\
Semantics             & 0.586 & 0.659   \\
\end{tabular}
\end{center}
\end{table}

\begin{figure*}[h]
%\vspace{.3in}
\centering
\includegraphics[scale=0.70]{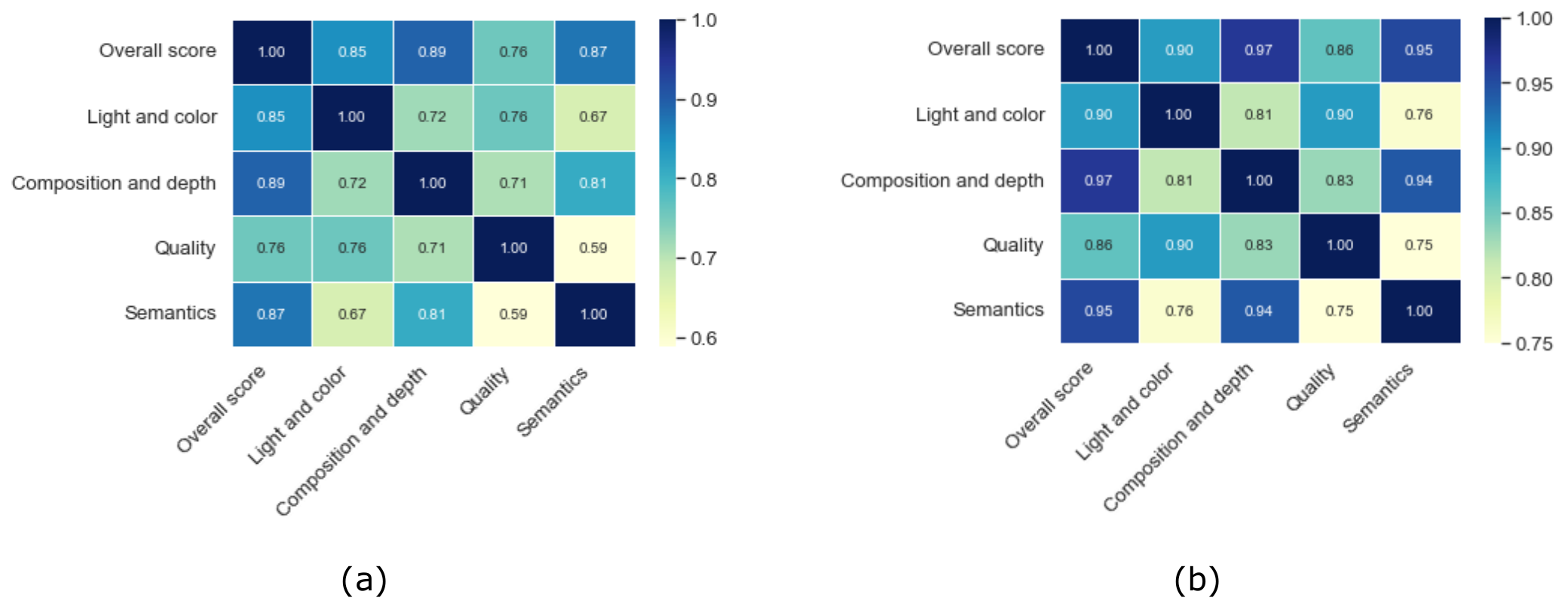}
%\vspace{.3in}
\caption{Spearman's rank correlations between the overall aesthetic scores and the attribute scores on the EVA dataset. \emph{(a)}: The ground-truth scores on all dataset,  \emph{(b)}: The predictions by our multi-task CNN on the test data.}
\label{correlations_eva}
\end{figure*}

Lastly, we report the Spearman's rank correlations on the EVA dataset in Figure~\ref{correlations_eva}. This figure shows the correlations for the ground-truth scores of all dataset on the left side, whereas on the right side, we present the correlations for the predictions on the test data made by our multi-task CNN. Our model's highest correlations among all attributes are for the composition and depth ($\rho$=0.97) and semantics ($\rho$=0.95) attributes. This finding is consistent with the EVA dataset, where composition and depth has the highest correlation with overall aesthetic scores ($\rho$=0.89) and semantics follows in second place ($\rho$=0.87). The lowest correlation belongs to the quality attribute, which is consistent with human data. Based on our evaluation, we can conclude that the predictions made by our multi-task CNN closely align with human interpretation in the EVA dataset as well.

\subsection{Cross-dataset evaluation}

In the final part of our analysis, we investigate the generalization capability of our multi-task CNN by conducting a cross-dataset evaluation. Firstly, we examine the performance of our model trained on the AADB dataset when tested on the test set of the EVA dataset. Subsequently, we reverse the process and evaluate the performance of our model trained on the EVA dataset when tested on the test set of the AADB dataset. The results of these cross-dataset evaluations are summarized in Table~\ref{crossdataset}. These results show the Spearman's rank correlations between the ground-truth overall aesthetic scores and the predictions made by our multi-task CNN.

\begin{table}[h]
\caption{Spearman's rank correlations between the ground-truth overall aesthetic scores and the predictions by our multi-task CNN for the cross dataset evaluation.} 
\label{crossdataset}
\begin{center}
\begin{tabular}{lcc}
\textbf{Train dataset} & \multicolumn{2}{c}{\textbf{Test dataset}} \\
\cline{1-3}
& \textbf{AADB \cite{kong2016}} & \textbf{EVA  \cite{kang2020}} \\
AADB & 0.707 & 0.321 \\
EVA & 0.441  & 0.695  \\
\end{tabular}
\end{center}
\end{table}

\newpage
Given the subjective nature of human aesthetics preferences, we acknowledge that the generalization of our multi-task CNN across datasets may be limited. However, interestingly, we find that when our model is trained on the EVA dataset and tested on the AADB dataset, it outperforms the vice versa scenario ($\rho$ = 0.441 $>$ $\rho$ = 0.321). One possible explanation for this observation is that the EVA dataset benefits from a larger number of ratings per image compared to the AADB dataset. This highlights the significance of high-quality human data and suggests that the availability of additional image aesthetic benchmarks could contribute to the development of models with improved generalization capabilities. Such models could offer valuable insights into understanding aesthetic preferences and uncovering underlying patterns associated with them.

\section{Conclusion}
\label{conclude}
Over the past decade, deep neural networks have achieved remarkable advancements in various domains, ranging from computer vision to game playing. Today, they have become an essential component of computational aesthetics. In this study, we present a simple yet effective multi-task CNN which simultaneously learns both the overall aesthetic scores and attribute scores of images. 

Through systematic evaluation, we have demonstrated the effectiveness of our neural network on two widely used image aesthetic benchmarks. Notably, our multi-task CNN surpasses existing approaches in the literature for the AADB dataset, establishing itself as the new state-of-the-art method for predicting overall aesthetic scores. Remarkably, our model achieves this superior performance while requiring fewer parameters compared to previous approaches. Furthermore, our study pioneers the application of a multi-task CNN on the EVA dataset, making it the first of its kind in this context.

Moving forward, we envision further advancements in computational aesthetics, facilitated by the integration of deep neural networks and the exploration of additional image aesthetic benchmarks. These developments have the potential to unlock deeper insights into the understanding and modeling of aesthetic preferences.

\appendices

\section*{Acknowledgment}
Funded by the European Union (ERC AdG, GRAPPA, 101053925, awarded to Johan Wagemans). Views and opinions expressed are however those of the authors only and do not necessarily reflect those of the European Union or the European Research Council Executive Agency. Neither the European Union nor the granting authority can be held responsible for them.

\enlargethispage{-7\baselineskip}
\begin{IEEEbiography}[{\includegraphics[width=1in,height=1.25in,clip,keepaspectratio]{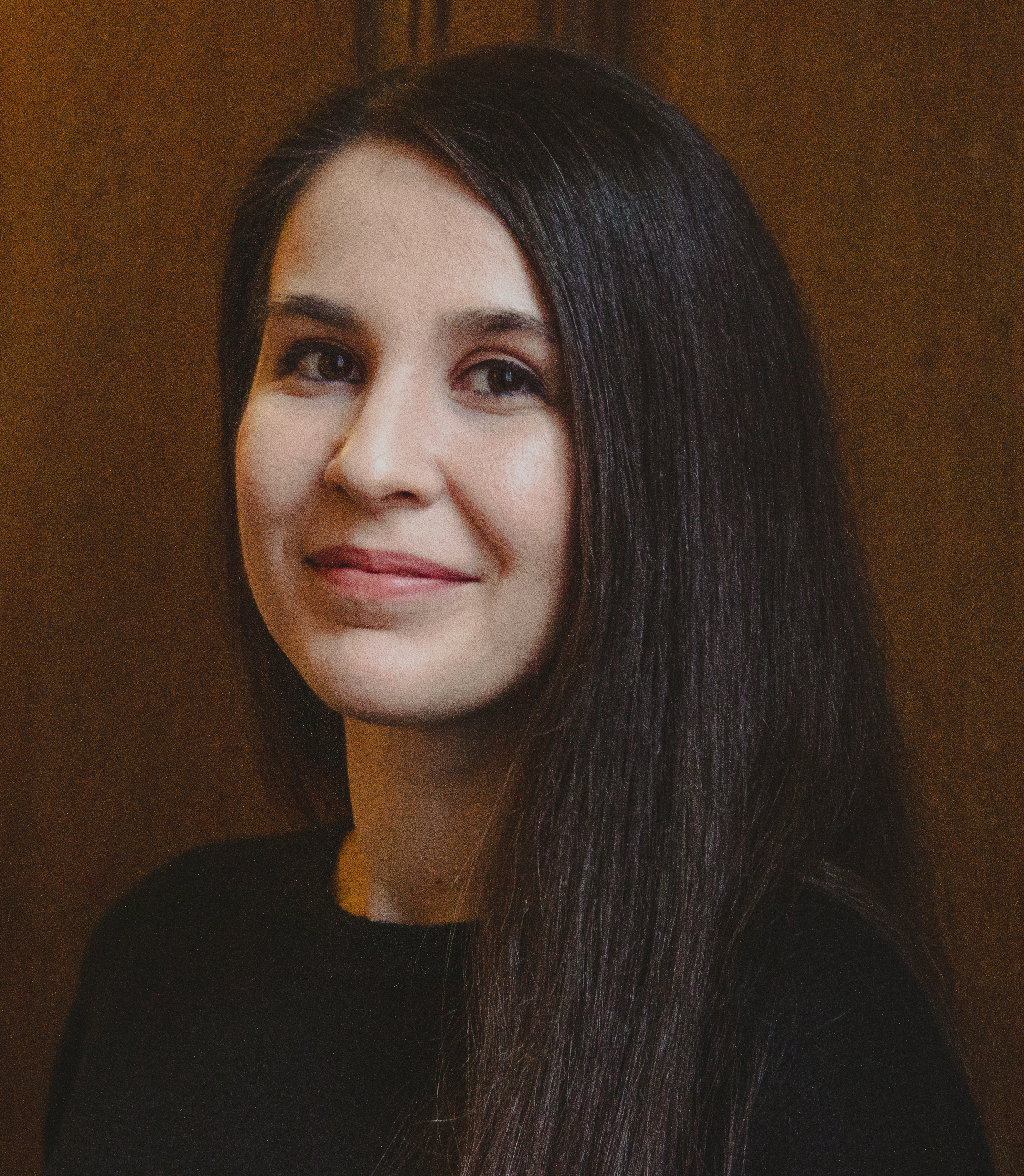}}]{Derya Soydaner} is a statistician and machine learning researcher focused on deep learning. She is currently a postdoctoral researcher in the Department of Brain \& Cognition at the University of Leuven (KU Leuven) and a member of the KU Leuven Institute for Artificial Intelligence. Her research interests include pattern recognition, machine learning and image processing.  
\end{IEEEbiography}

\vspace{-30\baselineskip}

\begin{IEEEbiography}[{\includegraphics[width=1in,height=1.25in,clip,keepaspectratio]{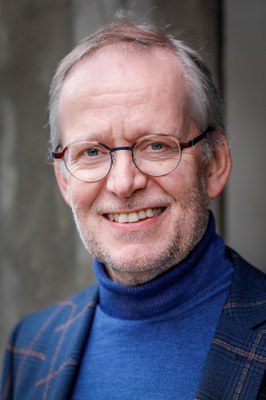}}]{Johan Wagemans} is Full Professor at the Department of Brain \& Cognition at the University of Leuven (KU Leuven) in Belgium. He has published more than 350 papers in international peer-reviewed journals, has edited the Oxford Handbook of Perceptual Organization, and is Senior Editor of Cognition, and Editor-in-Chief of Art \& Perception. He is currently leading two large interdisciplinary research programs on perception and appreciation of images and art, one funded by the Flemish Government (Methusalem) and one funded by the ERC (GRAPPA).
\end{IEEEbiography}

\EOD

\end{document}